\documentclass[conference]{IEEEtran}
\IEEEoverridecommandlockouts
\usepackage{cite}
\usepackage{amsmath,amssymb,amsfonts}
\usepackage[noend]{algorithmic}
\usepackage{algorithm}
\usepackage{graphicx}
\usepackage[table,xcdraw]{xcolor}
\usepackage{textcomp}
\usepackage{mwe}
\usepackage[normalem]{ulem}
\useunder{\uline}{\ul}{}
\usepackage{layouts}
\usepackage{comment}
\usepackage{multirow}
\usepackage{paralist}
\usepackage{url}
\usepackage{microtype}
\usepackage{calc}
\usepackage{booktabs}
\usepackage{xurl}
\usepackage{pifont}
\usepackage{mathrsfs}
\usepackage{subcaption}
\usepackage{hyperref}

\def\BibTeX{{\rm B\kern-.05em{\sc i\kern-.025em b}\kern-.08em
    T\kern-.1667em\lower.7ex\hbox{E}\kern-.125emX}}

\newcommand\thefontsize{The current font size is: \f@size pt}
\newcommand{\showFont}{encoding: \f@encoding{},
  family: \f@family{},
}
\newcommand\x{5px}
\newcommand\y{0px}
\DeclareRobustCommand{\saclass}{%
  \begingroup\normalfont
  \includegraphics[height=\fontcharht\font`\B+\y, width=\fontcharht\font`\B+\x]{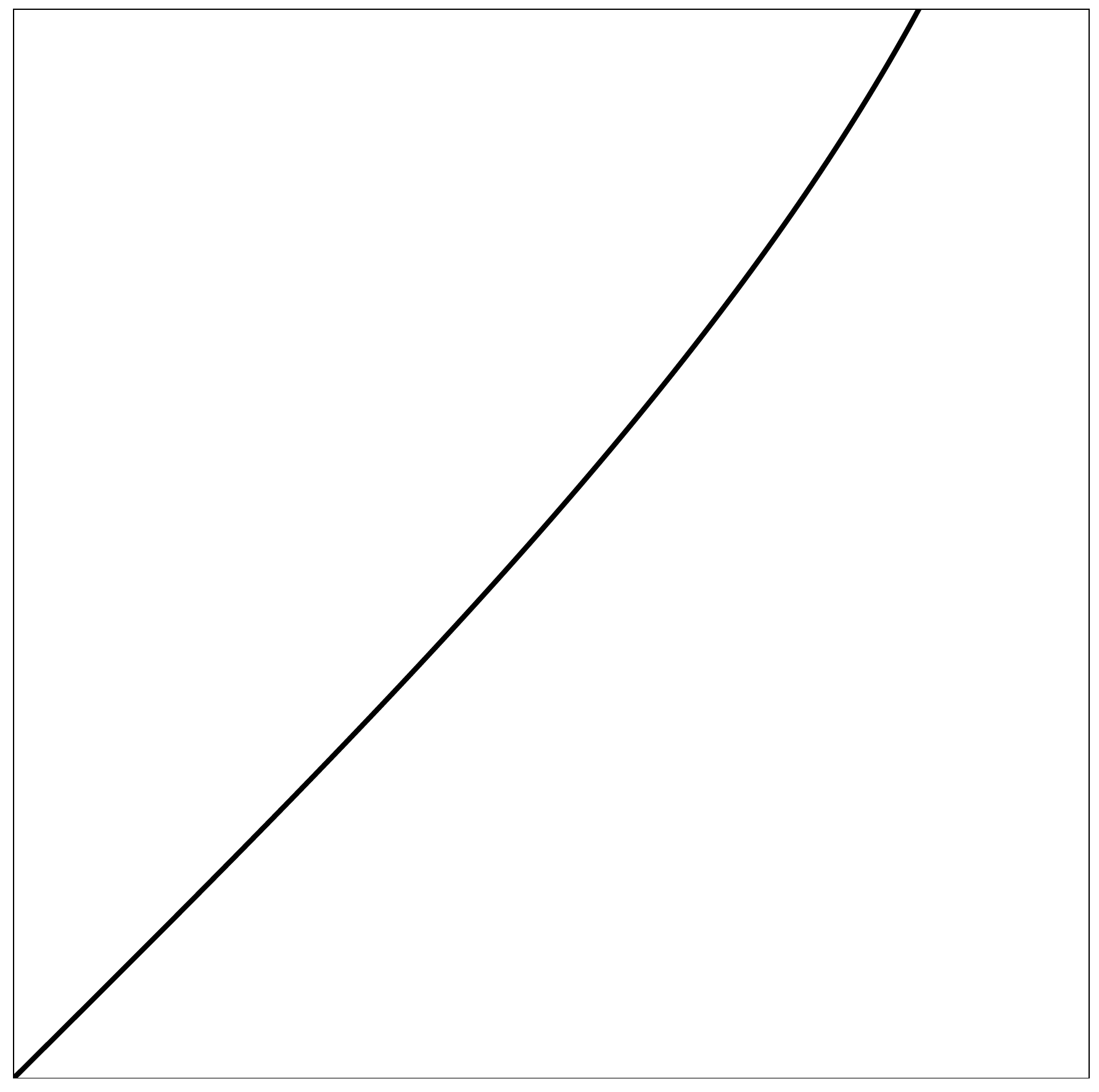}%
  \endgroup
}
\DeclareRobustCommand{\sbclass}{%
  \begingroup\normalfont
  \includegraphics[height=\fontcharht\font`\B+\y, width=\fontcharht\font`\B+\x]{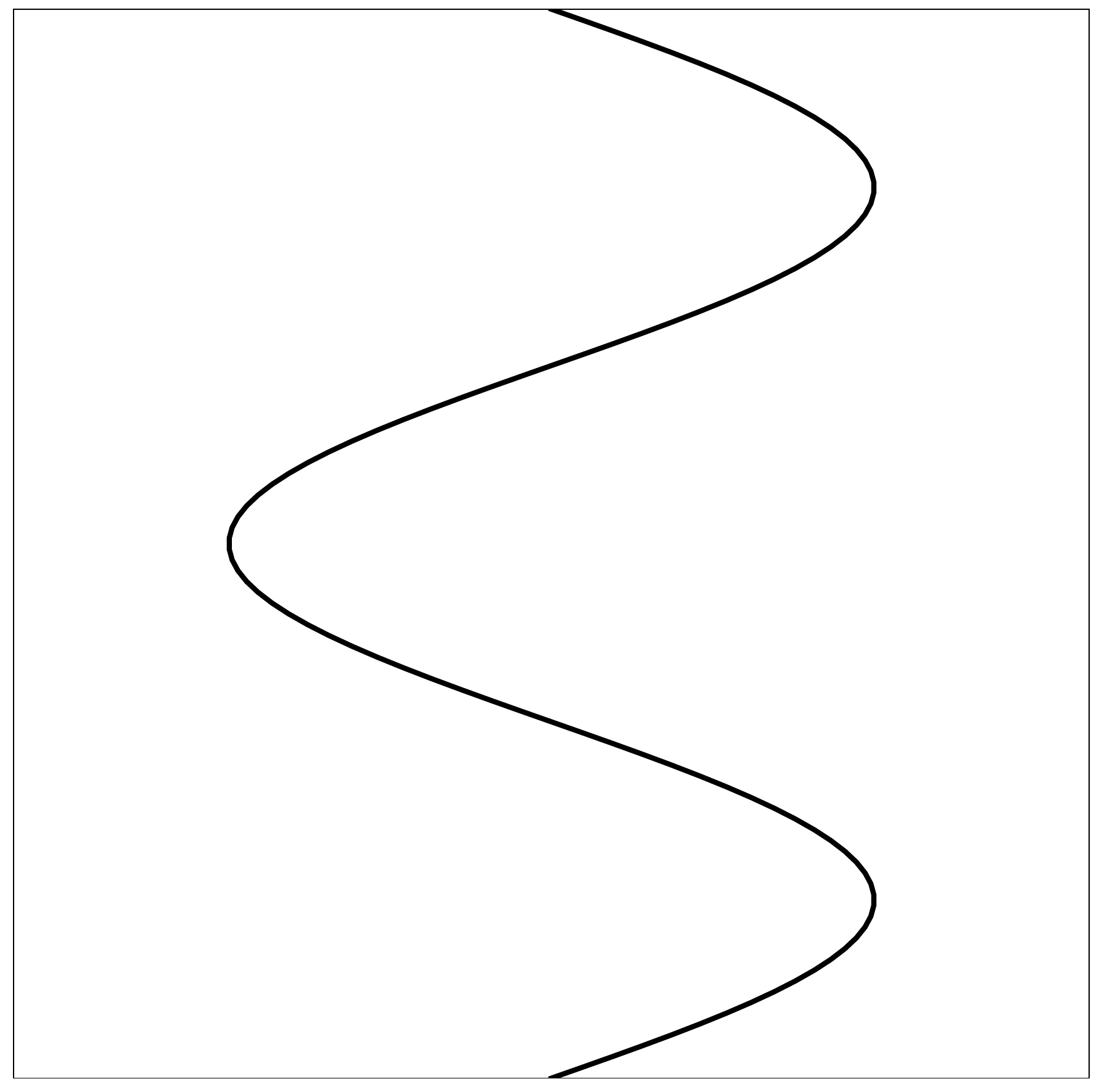}%
  \endgroup
}

\newcommand\blfootnote[1]{%
  \begingroup
  \renewcommand\thefootnote{}\footnotetext{#1}%
  \addtocounter{footnote}{-1}%
  \endgroup
}

\def\BibTeX{{\rm B\kern-.05em{\sc i\kern-.025em b}\kern-.08em
    T\kern-.1667em\lower.7ex\hbox{E}\kern-.125emX}}
\begin{document}

\title{MAcPNN: Mutual Assisted Learning on Data Streams with Temporal Dependence
}

\author{\IEEEauthorblockN{1\textsuperscript{st} Federico Giannini}
\IEEEauthorblockA{\textit{DEIB, Politecnico di Milano, Milano, Italy} \\
federico.giannini@polimi.it, \href{https://orcid.org/0000-0002-4210-6271}{ORCID}}
\and
\IEEEauthorblockN{2\textsuperscript{nd} Emanuele Della Valle}
\IEEEauthorblockA{\textit{DEIB, Politecnico di Milano, Milano, Italy} \\
emanuele.dellavalle@polimi.it, \href{https://orcid.org/0000-0002-5176-5885}{ORCID}}
}

\maketitle

\begin{abstract}
Internet of Things (IoT) Analytics often involves applying machine learning (ML) models on data streams. In such scenarios, traditional ML paradigms face obstacles related to continuous learning while dealing with concept drifts, temporal dependence, and avoiding forgetting. Moreover, in IoT, different edge devices build up a network. When learning models on those devices, connecting them could be useful in improving performance and reusing others' knowledge. This work proposes Mutual Assisted Learning, a learning paradigm grounded on Vygotsky's popular Sociocultural Theory of Cognitive Development. Each device is autonomous and does not need a central orchestrator. Whenever it degrades its performance due to a concept drift, it asks for assistance from others and decides whether their knowledge is useful for solving the new problem. This way, the number of connections is drastically reduced compared to the classical Federated Learning approaches, where the devices communicate at each training round. Every device is equipped with a Continuous Progressive Neural Network (cPNN) to handle the dynamic nature of data streams. We call this implementation Mutual Assisted cPNN (MAcPNN). To implement it, we allow cPNNs for single data point predictions and apply quantization to reduce the memory footprint. Experimental results prove the effectiveness of MAcPNN in boosting performance on synthetic and real data streams. 
\end{abstract}

\begin{IEEEkeywords}
data streams, concept drifts, temporal dependence, internet of things, assisted learning.
\end{IEEEkeywords}

\blfootnote{\copyright 2024 IEEE.
This is the author’s accepted manuscript of the paper:
Giannini, F., Della Valle, E. (2024). \textbf{MAcPNN: Mutual Assisted Learning on Data Streams with Temporal Dependence}. Proceedings of IEEE Big Data 2024, pp. 890–899.
The final authenticated version is available online at: \mbox{\url{https://doi.org/10.1109/BigData62323.2024.10825150}}}

\section{Introduction} \label{sec:introduction}
Within Internet of Things (IoT) applications, sensors consistently gather data, generating unbounded data flows known as data streams. When applying Machine Learning (ML) models in this scenario, reconsidering the paradigm is essential. Firstly, the knowledge is not entirely contained in a bounded dataset available at once, but data points are continuously generated over time. The challenge is to make predictions and update the model continuously as new data points arrive. Secondly, the hypothesis that data are independent and identically distributed (shortly i.i.d.) often does not hold. Data can, indeed, change its distribution and generate the problem known as \emph{concept drift}~\cite{cit:concept_drift}. ML algorithms should exhibit resilience to these changes, ensuring that performance remains consistent during and after the drift. Moreover, several works highlighted the importance of accounting for temporal dependence among data points to yield meaningful predictions~\cite{Ziffer22,cit:streaming_ts_zliobaite}. Finally, we do not want to lose the previous predictive ability when new distributions appear. This problem is known as \emph{catastrophic forgetting}~\cite{cit:catastrophic_forgetting}. While learning new knowledge, a model must achieve a trade-off between the ability to learn new information and the capacity to remember the past one~\cite{cit:stability_plasticity}. Even if, after a concept drift, we are usually interested in the current concept, remembering the past ones could be meaningful for many reasons. Old concepts may re-occur in the future. Additionally, new concepts can be similar to previous ones, or past knowledge can be useful in solving the current concept.

Usually, IoT applications rely on cloud computing, where data are collected by edge devices and transmitted to a centralized cloud for processing. However, this solution may have different disadvantages. Firstly, it may suffer latency that is incompatible with real-time requirements. Moreover, if edge devices produce a huge quantity of data, the available bandwidth may prove insufficient for transmitting all the data to the cloud. If the edge devices cannot process the data, a considerable portion may go uncaptured. These issues are particularly critical in a streaming scenario where data are continuously generated, and the learning model needs to process them promptly. Edge Computing~\cite{cit:edge_computing} steps in as a solution to these challenges. This paradigm moves the computation to edge devices, reducing latency and bandwidth costs, conserving battery life, and ensuring data safety and privacy. In an \emph{Edge ML} scenario, each edge device independently trains its model. Notably, the training procedure is also moved to the edges, not only the inference. The knowledge is, thus, distributed among different models. 

In an Edge ML scenario, connecting the different devices to allow them to share knowledge and improve themselves collaboratively may be crucial. When consolidating knowledge from multiple models, the main paradigm is \emph{Federated Learning (FL)}~\cite{cit:fedearated}. However, FL is primarily oriented towards federating knowledge from various devices to address a common general problem. It is not meant to build an approach where each device can leverage the knowledge of others to address its specific problem. FL aims to learn a shared model by aggregating locally computed updates in a network of models using a central node. To remove the need for a central orchestrator and build a network of peer-independent devices, \emph{Peer-to-peer FL (p2p FL)} proposed a scenario where each model independently computes its updates by aggregating the knowledge from its observational data and the one acquired by the others.

FL solutions cannot fit a streaming scenario that requires quick answers and adaptability. Consider, for instance, a network of weather stations spread across various geographical areas with different elevations, tasked with predicting the remaining useful life of different machinery (e.g., broadcast antennas, farm machinery) based on the weather data collected from a series of sensors. Each station has its own model and learns from its data. Some stations could be in remote areas with specific weather conditions that can affect the machinery's lives. Firstly, connecting the models at each training round could involve latency and bandwidth availability issues. The number of communications must be reduced. Additionally, each station can observe different concepts. Some concepts may be shared between the stations but can occur at different times. A concept could be a period with specific weather conditions, like drought or hard snow. The \emph{past} knowledge gained during another concept could be useful when learning a new concept. In this case, we are not interested in federating only the \emph{current} concept of the different stations. For these reasons, avoiding catastrophic forgetting in a network of devices is crucial. The knowledge associated with a past concept of a given device could be useful to assist another device in the future. Under these premises, we aim to investigate the following \textbf{research question}: \textit{is there a solution to connect a network of independent edge peer models continuously learning from data streams while minimizing inter-device communications?} Each edge ML model must be able to manage the challenges of concept drifts, temporal dependence, and catastrophic forgetting. We envision a paradigm where each device independently continuously trains its model and reuses the others' past knowledge to adapt to new concepts quickly.

We formulate a novel learning paradigm that we call \textbf{Mutual Assisted Learning (MAL)}. MAL is inspired by Vygotsky's popular Sociocultural Theory of Cognitive Development~\cite{cit:vygotsky}, highlighting the importance of social interaction during learning. This theory is based on the idea of \emph{Zone of Proximal Development (ZPD)}, which is the range of learning situated between what an individual can accomplish independently and what they can achieve with the assistance of another possessing knowledge and experience they do not yet possess. Learning becomes more effective when provided with appropriate assistance and guidance within the ZPD. In our network, individuals are the edge devices. When a device detects a concept drift, it may find itself in its ZPD and can ask for assistance from others to adapt to a new concept. It can then decide whether to use the assistance received from the others or continue the learning process on its own. This way, each device remains autonomous but can ask for the assistance of the others on demand. Moreover, it may improve others' learning process when asked for assistance. This bi-directional exchange of knowledge favours building rich and diversified shared knowledge. Integrating the on-demand mechanism significantly reduces communication, eliminating the need for continuous communication. Additionally, the adaptation can be boosted if a device has useful knowledge of a concept.

Our key contribution is \textbf{Mutual Assisted cPNNs (MAcPNN)} a novel approach that pioneers MAL in a network of \emph{Continuous Progressive Neural Networks (cPNN)}~\cite{cit:cpnn}. cPNN is our previous work that simultaneously addresses all the learning challenges from data streams. It sets at the confluence of three different research areas: Streaming Machine Learning (SML)~\cite{book_bifet}, Continual Learning (CL)~\cite{cit:cl}, and Time-Series Analysis (TSA)~\cite{cit:tsa}. 
However, cPNN is a periodic classifier requiring a mini-batch to make predictions. In this work, we thus introduce a \emph{new loss function} to optimize classification for each new data point. Furthermore, to favour edge devices with limited hardware and simplify the transfer of the models through a network, we reduce the memory footprint by using \emph{quantization}. Our experiments confirm that using MAL boosts performance on generated and real data streams.\footnote{All the code and data used in this work are available for reproducibility at \url{https://github.com/federicogiannini13/macpnn}}

The rest of the paper is organized as follows. Firstly, Section~\ref{sec:rel_works} analyzes the state of the art. Section~\ref{sec:method} exposes our method and contributions. Section~\ref{sec:exp} discusses the settings of our experiments, while Section~\ref{sec:results} exhibits the results. Finally, Section \ref{sec:conclusion} elaborates on conclusions, limits, and future works. 

\section{Related Works}\label{sec:rel_works}
A data stream is an unbounded sequence of data points \mbox{$D: d_0, d_1, ..., d_t, d_{t+1}, ...$} with $t \in \mathbb{N}$. In a data stream classification problem, each data point $d_t$ is a tuple $<X_t, y_t>$ where $X_t$ is the feature vector and $y_t$ is the associated label. The assumption is that, after receiving the feature vector $X_t$, the model must predict the associated label $\hat{y_t}$. The correct label $y_t$ will be available after the prediction and before receiving the new feature vector $X_{t+1}$. This way, one can apply a \emph{prequential evaluation} mechanism~\cite{cit:prequential} that, whenever a new data point $d_t$ is generated, acts as follows: 1) Infer the label $\hat{y_t}$ by inputting $X_t$ to the model. 2) Update a performance metric using the correct label $y_t$ 3) Update the model using $<X_t, y_t>$.

We call \emph{concept} the unobservable random process that produces the data points~\cite{cit:gama_learning_with_dd}. In this context, we should consider two main problems. Firstly, data can change its distribution over time. A concept drift is a phenomenon in which the statistical properties of a domain change over time in an arbitrary way~\cite{cit:concept_drift}. A concept drift is \emph{virtual} when there is a change in the probabilities $P(X|y)$ or $P(y)$, which notably does not alter the class boundary. In contrast, \emph{real concept drifts} impact the probability $P(y|X)$, thereby modifying the class boundary. Additionally, concept drifts can be classified based on the speed at which they happen. An \emph{abrupt} concept drift occurs instantly at time $t$, where the concept changes immediately from $t-1$ to $t$. In \emph{gradual} or \emph{incremental} concept drifts, the new concept progressively replaces the old one over time. Concepts can also recur over time. Secondly, data can exhibit temporal dependence. In this circumstance, given a data point $d_t$, $\exists \tau \; P(a_t|b_{t-\tau})\neq P(a_t)$ where $a_t \in X_t\cup \{y_t\}, b_{t-\tau} \in X_{t-\tau} \cup \{y_{t-\tau}\}$. This situation is crucial when it involves the label $y$, resulting in a correlation between the current label and the features or the labels of the previous data points.

\emph{Streaming Machine Learning (SML)}~\cite{book_bifet} focuses on learning continuously from single data points and detecting and managing concept drifts. It proposes streaming adaptations of decision trees and bagging. Two noteworthy examples are the Hoeffding Adaptive Trees (HAT)~\cite{DBLP:conf/sdm/BifetG07} and Adaptive Random Forests (ARF)~\cite{ARF}. They incorporate the ADaptive WINdowing (ADWIN) concept drift detector~\cite{ADWIN}. To address the temporal dependence, SML applies \emph{Temporal Augmentation (TA)}~\cite{cit:streaming_ts_zliobaite}. Denoting by $k$ the order of the TA, it adds to the feature space of each data point the labels of the previous $k$ data points ($X^{TA}_t = X_t \cup \{y_{t-1}, ..., y_{t-k}\}$).

\emph{Time Series Analysis (TSA)}~\cite{cit:tsa} offers methods to model sequences of observations collected over a specific time interval and arranged chronologically. Although it is historically based on statistical methods, in recent years, the forecasting potential of Deep Learning (DL) models has emerged~\cite{cit:makridakis_m5}. Variations of Long Short-Term Memory (LSTM)~\cite{cit:lstm} are widely applied.

\emph{Continual Learning (CL)}~\cite{cit:cl} addresses the forgetting problem in DL models. It assumes the data streams to be split into large batches of data called experiences. Data points in each experience are all accessible together in random order. It also assumes that each new experience introduces a drift. The main goal of a CL strategy is to incorporate into the model the knowledge coming from the new experience without forgetting the one associated with previous experiences.

In our previous work, we proposed \emph{Continuous Progressive Neural Networks (cPNN)}~\cite{cit:cpnn} to bridge CL, SML, and TSA. In the data stream classification problem, cPNN addresses the challenges of continuously learning from mini-batches of data points, managing temporal dependence, handling concept drifts, and preventing forgetting. It uses a continuous extension of LSTM (namely cLSTM and originally presented by Neto et al.~\cite{cit:ilstm}) to tame temporal dependence when learning continuously from a data stream. It accumulates data points in fixed-size mini-batches. When the mini-batch is complete, it builds sequences using a sliding window. Then, it trains a many-to-many LSTM model on the sequences. Given each sequence item, cLSTM outputs a score for each target class. The score $\hat{y}_c(d_t)$ for a given target class $c$ on data point $d_t$ is determined by averaging the scores $\hat{y}_c^s(d_t)$ from all sequences $s$ containing $d_t$ within the mini-batch. For each mini-batch, the loss function is computed as the Binary Cross Entropy, where the prediction for a given target class $c$ on data point $d_t$ is obtained as expressed in~(\ref{eq:cpnn_output}). cLSTM is, thus, a periodic classifier that can make predictions only on a mini-batch.
\begin{equation}\label{eq:cpnn_output}
\hat{y}_c(d_t) = Mean\{\hat{y}_c^{s}(d_t) | d_t \in s\}
\end{equation}

cPNN applies the CL strategy of \emph{Progressive Neural Networks (PNN)} to adapt to drifts quickly and avoid forgetting. It uses the PNN version named Gated Incremental Memory~\cite{cit:gim} to perform learning on sequences. When a concept drift occurs, it adds a new cLSTM (named \textit{column}) to the architecture. The feature vector of a given sequence item is built by concatenating the original feature vector with the output of the previous column's hidden layer associated with that item. This way, cPNN combines the knowledge acquired from the previous concepts with that directly acquired from the current one and boosts the adaptation. It is \emph{robust to forgetting} by construction since it freezes the parameters of the previous columns.

To the best of our knowledge, the literature currently does not contain works that address our research question. Although \emph{Federated Learning (FL)}~\cite{cit:fedearated} builds aggregated machine learning models using distributed data across multiple devices, its training procedure is orchestrated by a central node, and each training round requires communication between each node and the central node~\cite{cit:fl_survey}. 
Each device utilizes its local data for training and then transmits the model to the server for aggregation. The server subsequently disseminates the model update to participants, facilitating the achievement of the learning objective. More formally, it assumes to have a network of $n$ users' devices $\{U_1, ..., U_n\}$ with their own datasets $\{DS_1, ..., DS_n\}$ which may not be accessible by the other devices. Each training round is based on three steps: 1) The server sends the global model $M'$ model to the devices. 2) Each device $U_i$ trains its model $M_i$ by federating $DS_i$ and the model received by the server. 3) The server aggregates the models $\{M_1, ..., M_n\}$ to the new global model $M'$. Regarding FL applied to data streams, two main works explicitly address concept drifts~\cite{cit:fl_ds2,cit:fl_ds_fedstream} but do not tame temporal dependence between the single data points of a data stream. To overcome the need for a central orchestrator, some works~\cite{cit:p2p_fl1,cit:p2p_fl2,cit:p2p_fl3} proposed a \emph{Peer-to-peer FL (p2P FL)} framework. Each local model directly communicates with the others and updates its parameters. p2P FL requires peer communication at each training round, and there are no streaming applications yet.

\section{Proposed Method: a Pioneering Approach to Apply Mutual Assisted Learning in a Network of cPNNs} \label{sec:method}
In this section, we outline our contributions. First, Section~\ref{subsec:aclstm} introduces a new cLSTM loss function to overcome the need for mini-batches of data points, thus enabling anytime classifier scenarios. Next, Section~\ref{subsec:qcpnn} presents a novel solution that quantizes cPNN columns to reduce memory usage. Finally, in Section~\ref{subsec:fcpnn}, we demonstrate MAcPNN, a solution for applying MAL in a network of cPNNs.

\subsection{Anytime cLSTM}\label{subsec:aclstm}
The first limitation of cPNN and cLSTM is that they cannot provide classifications given a single data point. As Section~\ref{sec:rel_works} details, they require mini-batches of data points to perform training and inference. The classification of a data point $d_t$ is made by averaging all the predictions on $d_t$ associated with all the sequences containing $d_t$. Conversely, SML scenarios usually require an anytime classifier that makes predictions whenever a new data point arrives. Thus, we propose a new cLSTM loss function for the anytime classifier scenario. Traditional DL models typically run training on fixed-size mini-batches rather than individual data points. This approach facilitates the Stochastic Gradient Descent algorithm to produce more precise gradient estimates and enables parallel processing of data points~\cite{cit:deep_learning}. We preserve the training phase with mini-batches intact. However, we change the LSTM architecture and the loss function to align inference and training objectives. As for the original cLSTM, named $W$ the sliding window size and $B$ the mini-batch size, we build on each mini-batch $B-W+1$ sequences during the training phase. We change the LSTM architecture to a many-to-one and produce predictions only on the last item of each sequence. The new loss function is a Binary Cross Entropy that considers, for each mini-batch, the real and the predicted labels of each sequence's last item. The inference phase is performed whenever a new data point $d_t$ is generated. We, thus, consider the sequence containing the data points from $t-W+1$ to $t$. 

\subsection{Applying Quantization to Reduce cPNN's Memory Footprint}\label{subsec:qcpnn}
\begin{figure}
    \centering
    \begin{subfigure}{\linewidth}
    \centering
        \includegraphics[]{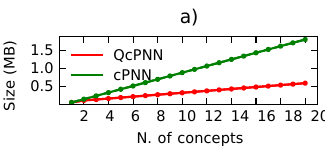}
        \phantomcaption{}
        \label{fig:quantization_1}
    \end{subfigure}

    \begin{subfigure}{\linewidth}
        \centering
        \includegraphics[]{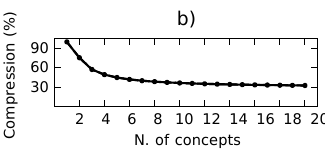}
        \phantomcaption{}
        \label{fig:quantization_2}
    \end{subfigure}
    
    \caption{Size of cPNN and QcPNN (\ref{fig:quantization_1}) and QcPNN Compression Ratio (\ref{fig:quantization_2}) considering data points with 10 features. Models are implemented using PyTorch, an LSTM hidden size of 50, and a window size of 10. INT8 quantization is applied. 
    }
    \label{fig:quantization}
\end{figure}

A second limit of cPNN is that its architecture grows linearly in the number of concepts. Each new concept adds a new column. The larger the number of concepts, the larger the memory footprint. Edge devices can be sensible to this problem since they may have limited hardware resources. To mitigate this problem, we propose Quantized cPNN (QcPNN). This novel version uses the quantization procedure to reduce the memory footprint of the frozen columns. Quantization is a method used to compute and save tensors with reduced data bandwidth requirements compared to floating-point precision~\cite{cit:quantization}.\footnote{PyTorch Quantization: \url{https://pytorch.org/docs/stable/quantization.html}} In a quantized model, computations are performed on tensors with lower precision, typically integer values instead of floating-point numbers. This process involves multiplying the floating-point values by a scale factor and rounding the result to obtain whole numbers when transitioning from floating-point to integer representation. The model's weights are quantized in advance in dynamic quantization, while activations are dynamically quantized during inference. This approach can be particularly advantageous for LSTM models, whose primary performance bottleneck is memory footprint rather than computational speed. When a new concept arises, QcPNN dynamically quantizes the last column before adding the new one and freezes its parameters. Fig.~\ref{fig:quantization} shows that, as the number of concepts increases, cPNN size in MB grows linearly with a high slope. The increase in the size of QcPNN is still linear but more contained. Its slope is between 25\% and 30\% of the cPNN's one. After five concepts, QcPNN halves the cPNN size. After ten concepts, QcPNN has a size equal to 35\% of cPNN's one.

\subsection{MAcPNN: Mutual Assisted cPNNs}\label{subsec:fcpnn}
\typeout{Page width: \the\textwidth}
Once we build cPNNs that can classify individual data points and reduce the memory footprint, we can propose a solution to connect a network of cPNNs using a MAL approach. We assume a network of $n$ users' devices with the associated data streams. Each data stream $DS_i$ observes different concepts. The same concept may appear on several data streams but at different timestamps. Data contain elaborated temporal dependence and concept drifts.

\begin{figure}[]
\centering
\includegraphics{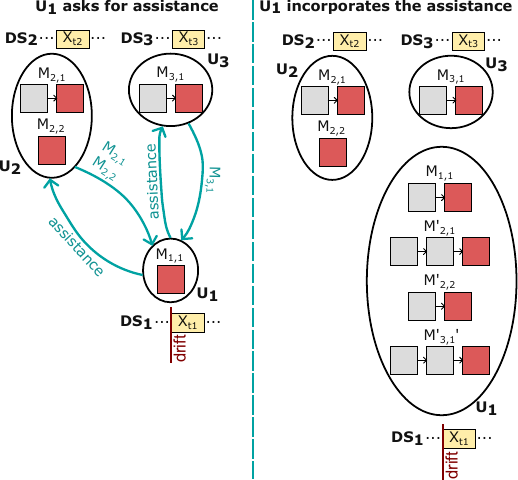}
\caption{MAL paradigm. When $U_1$ detects a drift on its data stream $DS_1$, it asks for assistance from the other devices. $U_2$ and $U_3$ send to $U_1$ copies of their local models. $U_1$ builds an ensemble containing its local models and copies of the others. $U_1$ adds a column to all the models. Grey represents the frozen columns, while red represents the trainable columns.} \label{fig:mal}
\end{figure}

We propose Mutual Assisted cPNN (MAcPNN) to address these challenges. The key idea of MAcPNN is to reuse local and external knowledge to boost the adaptation to new concepts following the theory of Vygotsky. Each device is autonomous and learns independently by training its QcPNN. When it encounters a concept drift, it asks for the assistance of others since it may find itself in its ZPD. In this case, using the knowledge of other peers may improve the learning process. After receiving assistance, it can decide if the assistance is helpful or if it can perform better by continuing its learning alone. More formally, each edge device $U_i$ starts with a QcPNN model trained on its associated data stream $DS_i$. We call a \emph{local model} of a device $U_i$ a QcPNN trained on $DS_i$ or the part of it. Upon encountering a concept drift, $U_i$ asks the others for copies of their local models. It then builds an ensemble comprising its local models and copies of others to decide whether the assistance is helpful. During a configurable number of training mini-batches, predictions are made by the best-performing model on the new concept. Then, the device selects the best-performing model. We call \emph{selection timestamp} the timestamp at which this selection occurs. Each device can store a maximum number of models to limit resource consumption. If the maximum is exceeded, $U_i$ selects subsets of the local and external models by considering a configurable proportion and favoring the models with higher selection timestamps. The MAL paradigm is represented in Fig.~\ref{fig:mal}.

Algorithm~\ref{pseudo:MAcPNN} details the training procedure of each device $U_i$. The output is a list \texttt{Pred} of predictions made on a data stream \texttt{DS$_i$}. $U_i$ stores a list \texttt{M} of models with the associated device ids \texttt{IDs} (where -1 stands for a local model) and selection timestamps \texttt{TS}. \texttt{M} initially contains only a QcPNN model. \texttt{E} represents the current concept's ensemble, while \texttt{Sel} represents the index of the currently selected model in the ensemble. The training procedure is expressed for each data point of \texttt{DS$_i$} represented by its features \texttt{X$_t$} and its target label \texttt{y$_t$}. When a concept drift is detected (Lines~\ref{pseudo:cd_start}-\ref{pseudo:cd_end}) if the ensemble contains more than one model, $U_i$ removes the last column from the models other than the best-performing one. It then updates the selected model's selection timestamp and node ID (set to -1 as a local model). At this point, it removes the external devices' models from \texttt{M}. Furthermore, it requests to each external device \texttt{u} its local models, appends them to \texttt{M}, and updates \texttt{IDs} and \texttt{TS} (Lines~\ref{pseudo:request_start}-\ref{pseudo:request_end}). After adding them to \texttt{M}, if the length of \texttt{M} exceeds \texttt{maxModels}, it keeps only \texttt{maxModels} models using \texttt{IDs}, \texttt{prop}, and \texttt{TS} (Lines~\ref{pseudo:max_start}-\ref{pseudo:max_end}). Then, it updates the ensemble \texttt{E} by taking the models of \texttt{M} and adding a column to each one (Lines~\ref{pseudo:ensemble_start}-\ref{pseudo:ensemble_end}). After the drift's check, $U_i$ can predict the label associated with the current data point using the best-performing model of \texttt{E} (Line~\ref{pseudo:pred}). Then, for each model in \texttt{E}, it updates performance and performs training (Lines~\ref{pseudo:preq_start}-\ref{pseudo:preq_end}). At this point, it can select the new best-performing model and update the counter \texttt{count} of data points of the current concept. $U_i$ updates the ensemble when \texttt{count} equals the number of data points contained in \texttt{numBatches} mini-batches. By indicating with \texttt{m} the current concept's best-performing model within the ensemble, it removes the last column from all the models other than \texttt{m}, it keeps in \texttt{E} only \texttt{m}, it sets to -1 the ID of \texttt{m}, it updates the selection timestamp of \texttt{m}, and it updates \texttt{M} (Lines~\ref{pseudo:selection_start}-\ref{pseudo:selection_end}).

\begin{algorithm}[t]
\caption{\mbox Training of user's edge device $U_i$ on the data stream $DS_i$}
\label{pseudo:MAcPNN}
\begin{algorithmic}[1]
\item[\textbf{Input:} \texttt{DS$_i$}, \texttt{numBatches}, \texttt{maxModels}, ]
\item[models' proportion \texttt{prop}, external devices \texttt{U$_{ext}$}]
\item[\textbf{Output:} Predictions \texttt{Pred}.]
\STATE \texttt{Pred} $\leftarrow [ ]$, \texttt{M} $\leftarrow [$QcPNN()$]$, \texttt{IDs} $\leftarrow$ $[$-1$]$, \texttt{TS} $\leftarrow$ $[$now()$]$
\STATE \texttt{Sel} $\leftarrow$ 0, \texttt{E} $\leftarrow$ \texttt{M}, \texttt{count} $\leftarrow$ 0, \texttt{Perf} $\leftarrow [$0$]$
\STATE \texttt{numDP} $\leftarrow$ \texttt{numBatches} * \texttt{M}$[$0$]$.batchSize
\FORALL{(\texttt{X$_t$}, \texttt{y$_t$}) in \texttt{DS$_i$}}
    \IF{conceptDrift()} \label{pseudo:cd_start}
        \IF{len(\texttt{E}) $>$ 1}
            \FORALL{\texttt{m} in \texttt{E} $\wedge$ \texttt{m} $\neq$ \texttt{E}$[$\texttt{Sel}$]$}
                \STATE \texttt{m}.removeLastColumn()
            \ENDFOR
            \STATE \texttt{TS}$[$\texttt{Sel}$]$ $\leftarrow$ now(), \texttt{IDs}$[$\texttt{Sel}$]$ $\leftarrow $ -1
            \STATE \texttt{M}, \texttt{IDs}, \texttt{TS} $\leftarrow$ removeExternalModels(\texttt{M}, \texttt{IDs}, \texttt{TS})
        \ENDIF
        \FORALL{\texttt{u} in \texttt{U$_{ext}$}} \label{pseudo:request_start}
            \STATE \texttt{M$_{ext}$} $\leftarrow$ \texttt{u}.getModels(), \texttt{M}.append(\texttt{M$_{ext}$}) 
            \FOR{\texttt{k} $\leftarrow$ 0; \texttt{k}$<$len(\texttt{M$_{ext}$}); \texttt{k} $\leftarrow$ \texttt{k}+1}
                \STATE \texttt{IDs}.append(\texttt{u}.ID), \texttt{TS}.append(\texttt{u}.TS$[$k$]$)
            \ENDFOR
            \IF{len(\texttt{M}) $>$ \texttt{maxModels}} \label{pseudo:max_start}
                \STATE \texttt{M}, \texttt{IDs}, \texttt{TS} $\leftarrow$ removeModels(\texttt{M}, \texttt{maxModels}, \texttt{prop}, \texttt{IDs}, \texttt{TS})
            \ENDIF \label{pseudo:max_end}
        \ENDFOR \label{pseudo:request_end}
        \STATE \texttt{count} $\leftarrow$ 0, \texttt{E} $\leftarrow$ \texttt{M}, \texttt{Sel} $\leftarrow$ 0, \texttt{Perf} $\leftarrow$ $[]$ \label{pseudo:ensemble_start}
        \FORALL{\texttt{m} in \texttt{E}}
            \STATE \texttt{m}.addNewColumn(), \texttt{Perf}.append(0)
        \ENDFOR \label{pseudo:ensemble_end}
    \ENDIF \label{pseudo:cd_end}
    \STATE \texttt{Pred}.append(\texttt{E}$[$\texttt{Sel}$]$.predict($X_t$))  \label{pseudo:pred} 
    \FOR{\texttt{k} $\leftarrow$ 0; \texttt{k}$<$len(\texttt{E}); \texttt{k} $\leftarrow$ \texttt{k}+1} \label{pseudo:preq_start}
            \STATE \texttt{Perf}$[$\texttt{k}$] \leftarrow$ updatePerf(\texttt{Perf}$[$\texttt{k}$]$, \texttt{E}$[$\texttt{k}$]$.predict(\texttt{X$_t$}), \texttt{y$_t$}), \texttt{E}$[$\texttt{k}$]$.train(\texttt{X$_t$}, \texttt{y$_t$})
    \ENDFOR \label{pseudo:preq_end}
    \STATE \texttt{count} $\leftarrow$ \texttt{count}+1, \texttt{Sel} $\leftarrow$ argMax(\texttt{Perf})
    \IF{\texttt{count} = \texttt{numDP}} \label{pseudo:selection_start}
        \FORALL{\texttt{m} in \texttt{E} $\wedge$ \texttt{m} $\neq$ \texttt{E}$[$\texttt{Sel}$]$}
            \STATE \texttt{m}.removeLastColumn()
        \ENDFOR
        \STATE \texttt{E} $\leftarrow$ $[$\texttt{E}$[$\texttt{Sel}$]]$
        \STATE \texttt{Perf} $\leftarrow$ $[$\texttt{Perf}$[$\texttt{Sel}$]]$, \texttt{TS}$[$\texttt{Sel}$]$ $\leftarrow $ now()
        \STATE \texttt{IDs}$[$\texttt{Sel}$]$ $\leftarrow $ -1, \texttt{Sel} $\leftarrow $0
        \STATE \texttt{M}, \texttt{IDs}, \texttt{TS} $\leftarrow$ removeExternalModels(\texttt{M}, \texttt{IDs}, \texttt{TS})
    \ENDIF \label{pseudo:selection_end}
\ENDFOR
\end{algorithmic}
\end{algorithm}

Notably, each device conducts the entire process autonomously without a central orchestrator. Communication between the devices is minimized by asking for assistance only when a concept drift is detected. Using QcPNN reduces the memory footprint, making it easier to transfer the models and apply an ensemble. Given a period where each device detects one drift and assuming that all the devices receive the same number of training mini-batches $N_B$, we can quantify the total number of communications over the network with~(\ref{eq:communications}), where $n$ is the number of devices. $C_{naive}$ represents a naive approach in which each device communicates with the other $n-1$ at each training round. $C_{ours}$ represents our approach in which each device communicates with the other $(n-1)$ only when it detects a drift. Since each device detects one drift, the total number of drifts in the network is $n$. $C_{ours}$ reduces $C_{naive}$ by $1/N_B$. Considering the entire data streams, assuming that drifts occur on average after $B_D$ mini-batches on each data stream, the total number of communications is \mbox{$n(n-1)/B_D$}. Additionally, to reduce the number of communications, each device could communicate only with the devices in its neighborhood. 

\begin{equation}
\label{eq:communications}
C_{naive}=n(n-1)N_B,\;\;\;\;C_{ours} = n(n-1)
\end{equation}

\section{Experimental Environment Setup} \label{sec:exp}
In this section, we provide an overview of the setup of the experiments. As reported in cPNN's paper~\cite{cit:cpnn}, the most commonly used SML benchmarks are unsuitable for our purpose or do not consider temporal dependence. Therefore, we build a new benchmark.  Section~\ref{subsec:datastreams} explains the data streams' generation. Section~\ref{subsec:hypothesis} covers hypothesis formulation and the evaluation methodology.

\subsection{Generated Data Streams}\label{subsec:datastreams}
We start from the \emph{SineRW} generator (SRW) presented in~\cite{cit:cpnn} that generates points in two dimensions ($x_1$ and $x_2$) with a random walk procedure and classifies them using two boundary functions. We generalize them and produce more boundary functions by introducing $S_1$ and $S_2$ (\ref{eq:sine1_generalized}). $S_1$ serves as \emph{simple} boundary functions, whereas $S_2$ serves as \emph{complex} boundary functions. $S_1$ is, in fact, just a curve in two dimensions (i.e., \saclass{}), while $S_2$ pulsates twice (i.e., \sbclass{}). Each boundary function yields two binary classification functions: the first assigns 1 to points meeting the condition obtained by replacing the equality with the inequality $\geq 0$, while the others are assigned 0. The second classification function uses  $< 0$.

\begin{equation}
\label{eq:sine1_generalized}
S_1: x_1 - \alpha - \beta sin(\gamma x_2) = 0,\;\;S_2:x_1 - \alpha - \beta sin(\gamma \pi x_2)=0
\end{equation}

We then introduce an elaborated temporal dependence. The new label at time $t$, denoted as $y'_{t}$, is calculated using~(\ref{eq:sine_rw_mode}), where $y_t$ is the binary label assigned by the original SRW. We rely on current and previous labels.
\begin{equation}
\label{eq:sine_rw_mode}
y_t' = MODE(y_t, y_{t-1}, y_{t-2}, y_{t-3}, y_{t-4})
\end{equation}

Moreover, we use \emph{Weather}~\cite{cit:weather} and \emph{AirQuality}\footnote{\url{https://data.seoul.go.kr/dataList/OA-15526/S/1/datasetView.do}} datasets to prove our hypothesis in realistic scenarios. Weather includes two stations' continuous hourly hydrometeorological variables (air temperature, relative humidity, wind speed, wind direction, dew point temperature), resulting in 96432 data points each. AirQuality includes continuous hourly detections from 2017 to 2020 of 4 gases (SO2, NO2, O3, CO) and particulate matter (PM2.5 and PM10) from 25 stations located in Seoul. Each station contains about 34k data points. We select a target feature $v$ to construct binary labels for a classification problem (the air temperature for Weather, NO2 for AirQuality). We then create five binary classification functions as expressed in ~(\ref{eq:f1+})-(\ref{eq:f5+}) to incorporate elaborate temporal dependence. Here, $v_t$ is the value of the target feature of the data point at timestamp $t$, and $k$ is the temporal order (10 for Weather, 5 for AirQuality). We express the increment at time $t$ as \mbox{$\Delta_t = v_t-v_{t-1}$}. $Med$ represents the median, while $Min$ the minimum. The rationale is to produce binary labels with a maximum unbalance percentage of 70\%-30\%. Labels assigned by each function must differ from more than 20\% to those assigned by the others. We aim to build a complex problem in which the current label often changes concerning the previous. This way, a simple model that predicts the current label as the previous one does not perform well. Ultimately, the target feature is removed from the feature vector. We can imagine five boundary functions. $F_1$, $F_2$, and $F_3$ compare the current value with the previous one, the median of the previous $k$, and the minimum of the previous $k$, respectively. $F_4$ and $F_5$ compare the current increment with the previous one and the median of the previous $k$. Each boundary function produces two classification functions. The negative classification functions $F_{1-}$, ..., $F_{5-}$ invert labels of positive ones reported in~(\ref{eq:f1+})-(\ref{eq:f5+}).
\begin{equation}
\label{eq:f1+}
\begin{split}
  F_{1+}: y(X_t) = 
  \begin{cases}
    1, & \text{if } v_t > v_{t-1} \\
    0, & \text{otherwise}
  \end{cases}
\end{split}
\end{equation}
\begin{equation}
\label{eq:f2+}
  \begin{aligned}
    F_{2+}: y(X_t) = 
    \begin{cases}
      1, & \text{if } v_t > Med(v_{t-k}, \ldots, v_{t-1}) \\
      0, & \text{otherwise}
    \end{cases}
  \end{aligned}
\end{equation}
\begin{equation}
\label{eq:f3+}
  F_{3+}: y(X_t) = 
  \begin{cases}
    1, & \text{if } v_t > Min(v_{t-k}, ..., v_{t-1})\\
    0, & \text{otherwise}
  \end{cases}
\end{equation}
\begin{equation}
\label{eq:f4+}
  F_{4+}: y(X_t) = 
  \begin{cases}
    1, & \text{if } \Delta_t > \Delta_{t-1}\\
    0, & \text{otherwise}
  \end{cases}
\end{equation}
\begin{equation}
\label{eq:f5+}
  F_{5+}: y(X_t) = 
  \begin{cases}
    1, & \text{if } \Delta_t > Med(\Delta_{t-k}, ..., \Delta_{t-1})\\
    0, & \text{otherwise}
  \end{cases}
\end{equation}

We simulate a three-device ($U_1$, $U_2$, $U_3$) network, each associated with a data stream with five concepts. Each SRW concept contains 25k data points. 
To reduce the intersection between the data streams, we split Weather data from each station into four timestamp-based segments, minimizing data overlap across streams. $U_1$ observes the segments of the first station and the first segment of the second station. $U_2$ observes the first station's last three segments and the second station's first two segments. $U_3$ observes the last segment of the first station and all segments of the second station. Each segment represents a concept. For AirQuality, we select the 15 stations with the highest number of data points and assign five different stations to each device. Each station represents a concept.

\begin{figure}[t]
\centering
\includegraphics{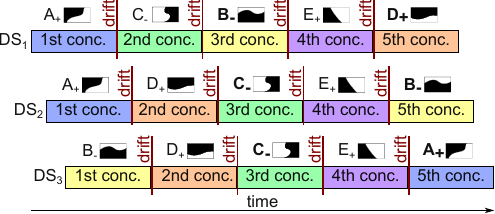}
\caption{Concept organization within the data streams of three users' devices. We combine five classification functions with abrupt drifts: some concepts are seen by others, and some are new. We boldly highlight concepts seen by other devices. Drifts are unsynchronized across streams.} \label{fig:concepts}
\end{figure}

We assign a specific classification function to each concept and organize them on the different data streams, as shown in Fig.~\ref{fig:concepts}. We obtain an organization in which some concepts introduce new classification functions while others (the third and fifth) introduce already-seen functions in the other devices. Concept drifts are abrupt and not synchronous. Indeed, the devices are run in parallel but start at different timestamps. $U_2$ and $U_3$ start respectively after thousands of data points of $U_1$ and $U_2$ (2k for SRW and Weather, 3k for AirQuality).  The different configurations for SRW are produced by assigning different combinations of $S_1$ and $S_2$ functions to A, B, C, D, and E. Particularly, we consider 16 $S_1$ boundary functions with $\alpha \in \{0, 1\}, \beta \in \{-1, 1\}, \gamma \in [0.8, 1.2]$. We also consider 16 $S_2$ boundary functions with $\alpha = 0.5, \beta \in [-0.25, -0.15]$, \mbox{$\gamma \in [-2.2, -1.8 ]$}. Each configuration contains a maximum of three $S_1$ or $S_2$ functions. For Weather and AirQuality, we produce different configurations by assigning to A, B, C, D, and E different permutations of $F_1$, $F_2$, $F_3$, $F_4$, and $F_5$. 

\subsection{Hypothesis Formulation}\label{subsec:hypothesis}
Our \textbf{hypothesis} is the following: \textit{in a three-user device network, MAcPNN outperforms SML models, cPNN, and cLSTM trained solely on the device-specific data streams $DS_i$, considering the average performance across the entire network}.

We compare the following models for each device: ARF, ARF$_T$, HAT, HAT$_T$, cLSTM (see Sections~\ref{sec:rel_works} and \ref{subsec:aclstm}), cPNN and MAcPNN. HAT$_T$ and ARF$_T$ represent HAT and ARF versions with TA. All the models except MAcPNN are trained only on the data stream associated with the specific device. MAcPNN can reuse the knowledge gained from other devices. As stressed by Section~\ref{sec:rel_works}, FL solutions are not considered since they do not fit our problem's requirements. Our hypothesis tests whether the mutual assistance between the devices made by MAcPNN improves performance concerning the models trained only on the specific device's data. Since the labels are not perfectly balanced, we consider \emph{Cohen's Kappa} score~\cite{cit:cohen_kappa} and \emph{Balanced Accuracy} as the evaluation metrics.  We apply the \emph{prequential evaluation} (which tests each data point's prediction before training on it) for each data stream concept. The performance at data point $d_t$ is calculated from the first data point following the last drift to $d_t$. We highlight two specific moments for each concept originated by the drift $j$: its start (\emph{start$_j$}) and its completion (\emph{end$_j$}). \emph{start$_j$} represents the first 50 learning mini-batches ($50*128$ data points).\footnote{The mini-batch size is 128. The first good performance was reached in 50 mini-batches.} It is useful to measure how the model reacts to the drift $j$. \emph{end$_j$} is the score computed on the last data point of the concept originated by the drift $j$. It is, thus, associated with all the data points belonging to the concept originated by the drift $j$. It is useful to measure how the model performs on the whole concept. We exclude the initial concept (the one before the first drift) since cLSTM, cPNN, and MAcPNN share identical architectures. Given a device, we average the different start$_j$ and end$_j$ performance on all the concept drifts. Finally, we average the results of the three devices and produce start$_{avg}$ and end$_{avg}$. This way, the start$_{avg}$ performance represents the average performance of the three devices after the first 50 mini-batches of the different concepts. The end$_{avg}$ performance signifies the average performance after the end of the concepts on the three devices. We conduct experiments on generated and real data streams, including SRW, Weather, and AirQuality configurations. Additionally, we assume to have a concept drift detector with 100\% accuracy. 

After executing the preliminary experiments, hyperparameter values are selected as follows (similarly to the cPNN's paper~\cite{cit:cpnn}). Epochs number 10; mini-batch size 128; learning rate: 0.01; LSTM hidden layer size: 50; LSTM Window Size: 10 for SRW and AirQuality, 11 for Weather. INT8 quantization is used for QcPNN. We use the default settings for the SML models and refer to the LSTM window size to set the TA order. For MAcPNN, we use the following parameters. \texttt{maxModels}: 10, \texttt{prop}: 0.3, \texttt{numBatches}: 50. cLSTM, cPNN and MAcPNN models are initialized in the same way on all the devices to reduce the variability of results because of the initialization.

\section{Results} \label{sec:results}

\begin{figure*}[]
\centering
\includegraphics{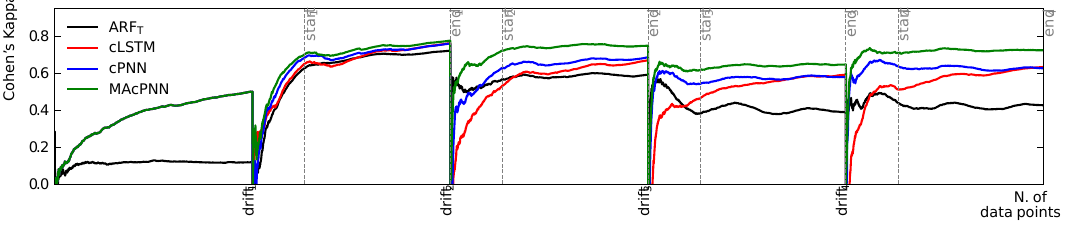}
\caption{An example of Cohen's Kappa evolution over time of Device 2 on a Weather configuration to highlight the points in which we observe \emph{start} and \emph{end} metrics. Scores are reset after each drift. The score at data point $d_t$ considers the predictions from the first data point following the last drift to $d_t$.}  \label{fig:performance_node}
\end{figure*}

 \begin{table*}\caption{\textbf{Cohen's Kappa} average scores and the corresponding standard deviations for 10 SRW configurations and 50 Weather (W) and AirQuality (AQ) configurations. Given a configuration, for each concept drift $j$, we compute the mean values of start$_j$ and end$_j$ over the three devices of the network. start$\mathbf{_{avg}}$ and end$\mathbf{_{avg}}$ are computed by first averaging the different start$_j$ and end$_j$ values for each device separately and then averaging these three values. We then compute the mean and the standard deviation over all the configurations of each type of data stream (SRW, W, and AQ). We highlight the statistically best-performing model in bold. MAcPNN is always the best model after the beginning of the concepts. After the concepts end, cLSTM can approach MAcPNN only on Weather. 
 }\label{table:results_kappa}
\centering

\begin{tabular}{cc|cccc|c|cccc|c|}
\cline{3-12}
\textbf{}                                                                               & \textbf{} & \textbf{start$\mathbf{_1}$}  & \textbf{start$\mathbf{_2}$}  & \textbf{start$\mathbf{_3}$}  & \textbf{start$\mathbf{_4}$}  & \textbf{start$\mathbf{_{avg}}$} & \textbf{end$\mathbf{_1}$}    & \textbf{end$\mathbf{_2}$}    & \textbf{end$\mathbf{_3}$}    & \textbf{end$\mathbf{_4}$}    & \textbf{end$\mathbf{_{avg}}$}  \\ \hline
\multicolumn{1}{r|}{\multirow{4}{*}{\begin{tabular}[c]{@{}r@{}}S\\ R\\ W\end{tabular}}} & ARF$_T$   & 0.74±0.00          & 0.74±0.00          & 0.73±0.00          & 0.73±0.00          & 0.73±0.00           & 0.73±0.00          & 0.73±0.00          & 0.73±0.00          & 0.73±0.00          & 0.73±0.00          \\
\multicolumn{1}{r|}{}                                                                   & cLSTM     & 0.69±0.05          & 0.69±0.05          & 0.65±0.06          & 0.73±0.05          & 0.69±0.03           & 0.80±0.04          & 0.81±0.04          & 0.78±0.06          & 0.84±0.03          & 0.81±0.02          \\
\multicolumn{1}{r|}{}                                                                   & cPNN      & 0.67±0.03          & 0.72±0.06          & 0.75±0.07          & 0.71±0.02          & 0.71±0.02           & 0.80±0.02          & 0.81±0.04          & 0.81±0.06          & 0.78±0.02          & 0.80±0.02          \\
\multicolumn{1}{r|}{}                                                                   & MAcPNN    & 0.74±0.06          & \textbf{0.84±0.06} & \textbf{0.80±0.06} & \textbf{0.85±0.04} & \textbf{0.81±0.03}  & \textbf{0.83±0.03} & \textbf{0.88±0.04} & 0.85±0.06          & \textbf{0.88±0.03} & \textbf{0.86±0.02} \\ \hline
\multicolumn{1}{r|}{\multirow{4}{*}{W}}                                                 & ARF$_T$   & 0.35±0.12          & 0.42±0.15          & 0.43±0.21          & 0.43±0.08          & 0.41±0.04           & 0.40±0.12          & 0.45±0.15          & 0.44±0.21          & 0.46±0.07          & 0.44±0.04          \\
\multicolumn{1}{r|}{}                                                                   & cLSTM     & 0.47±0.06          & 0.55±0.09          & 0.48±0.13          & 0.62±0.06          & 0.53±0.03           & 0.62±0.05          & 0.67±0.07          & 0.62±0.11          & 0.70±0.04          & 0.65±0.02          \\
\multicolumn{1}{r|}{}                                                                   & cPNN      & 0.57±0.05          & 0.59±0.08          & 0.57±0.12          & 0.56±0.05          & 0.57±0.03           & 0.63±0.06          & 0.63±0.08          & 0.58±0.12          & 0.57±0.05          & 0.60±0.03          \\
\multicolumn{1}{r|}{}                                                                   & MAcPNN    & \textbf{0.59±0.05} & \textbf{0.63±0.07} & \textbf{0.62±0.10} & \textbf{0.67±0.04} & \textbf{0.63±0.02}  & 0.64±0.05          & 0.66±0.08          & \textbf{0.65±0.10} & 0.69±0.04          & 0.66±0.02          \\ \hline
\multicolumn{1}{r|}{\multirow{4}{*}{\begin{tabular}[c]{@{}r@{}}A\\ Q\end{tabular}}}     & ARF$_T$   & 0.21±0.09          & 0.21±0.11          & 0.19±0.13          & 0.19±0.06          & 0.20±0.03           & 0.26±0.10          & 0.27±0.11          & 0.24±0.14          & 0.25±0.06          & 0.26±0.03          \\
\multicolumn{1}{r|}{}                                                                   & cLSTM     & 0.35±0.04          & 0.34±0.05          & 0.29±0.05          & 0.32±0.03          & 0.32±0.02           & 0.41±0.04          & 0.38±0.05          & 0.36±0.05          & 0.36±0.03          & 0.38±0.02          \\
\multicolumn{1}{r|}{}                                                                   & cPNN      & 0.40±0.03          & 0.39±0.05          & 0.35±0.05          & 0.33±0.03          & 0.37±0.02           & 0.41±0.04          & 0.38±0.05          & 0.35±0.05          & 0.33±0.03          & 0.37±0.02          \\
\multicolumn{1}{r|}{}                                                                   & MAcPNN    & \textbf{0.43±0.03} & \textbf{0.43±0.05} & \textbf{0.41±0.04} & \textbf{0.40±0.03} & \textbf{0.42±0.02}  & \textbf{0.43±0.04} & \textbf{0.41±0.05} & \textbf{0.41±0.04} & \textbf{0.40±0.03} & \textbf{0.41±0.02} \\ \hline
\end{tabular}
 
\end{table*}
 \begin{table*}\caption{\textbf{Balanced Accuracy} average scores and the corresponding standard deviations for 10 SRW configurations and 50 Weather (W) and AirQuality (AQ) configurations. MAcPNN is always the best model after the beginning of the concepts. After the concepts end, cLSTM can approach MAcPNN only on Weather. 
 }\label{table:results_acc}
\centering

\begin{tabular}{cc|cccc|c|cccc|c|}
\cline{3-12}
\textbf{}                                                                               & \textbf{} & \multicolumn{1}{c}{\textbf{start$\mathbf{_1}$}} & \multicolumn{1}{c}{\textbf{start$\mathbf{_2}$}} & \multicolumn{1}{c}{\textbf{start$\mathbf{_3}$}} & \multicolumn{1}{c|}{\textbf{start$\mathbf{_4}$}} & \multicolumn{1}{c|}{\textbf{start$\mathbf{_{avg}}$}} & \multicolumn{1}{c}{\textbf{end$\mathbf{_1}$}} & \multicolumn{1}{c}{\textbf{end$\mathbf{_2}$}} & \multicolumn{1}{c}{\textbf{end$\mathbf{_3}$}} & \multicolumn{1}{c|}{\textbf{end$\mathbf{_4}$}} & \multicolumn{1}{c|}{\textbf{end$\mathbf{_{avg}}$}} \\ \hline
\multicolumn{1}{r|}{\multirow{4}{*}{\begin{tabular}[c]{@{}r@{}}S\\ R\\ W\end{tabular}}} & ARF$_T$   & 0.87±0.00                             & 0.87±0.00                             & 0.87±0.00                             & 0.87±0.00                              & 0.87±0.00                                & 0.87±0.00                           & 0.87±0.00                           & 0.87±0.00                           & 0.87±0.00                            & 0.87±0.00                              \\
\multicolumn{1}{r|}{}                                                                   & cLSTM     & 0.84±0.02                             & 0.84±0.02                             & 0.82±0.03                             & 0.87±0.02                              & 0.84±0.02                                & 0.90±0.02                           & 0.90±0.02                           & 0.89±0.03                           & 0.92±0.01                            & 0.90±0.01                              \\
\multicolumn{1}{r|}{}                                                                   & cPNN      & 0.83±0.02                             & 0.86±0.03                             & 0.87±0.04                             & 0.85±0.01                              & 0.86±0.01                                & 0.90±0.01                           & 0.91±0.02                           & 0.91±0.03                           & 0.89±0.01                            & 0.90±0.01                              \\
\multicolumn{1}{r|}{}                                                                   & MAcPNN    & 0.87±0.03                             & \textbf{0.92±0.03}                    & \textbf{0.90±0.03}                    & \textbf{0.93±0.02}                     & \textbf{0.90±0.01}                       & \textbf{0.92±0.02}                  & \textbf{0.94±0.02}                  & 0.92±0.03                           & \textbf{0.94±0.02}                   & \textbf{0.93±0.01}                     \\ \hline
\multicolumn{1}{r|}{\multirow{4}{*}{W}}                                                 & ARF$_T$   & 0.68±0.06                             & 0.71±0.08                             & 0.71±0.11                             & 0.72±0.04                              & 0.70±0.02                                & 0.70±0.06                           & 0.72±0.07                           & 0.72±0.11                           & 0.73±0.04                            & 0.72±0.02                              \\
\multicolumn{1}{r|}{}                                                                   & cLSTM     & 0.73±0.03                             & 0.77±0.05                             & 0.74±0.07                             & 0.81±0.03                              & 0.76±0.01                                & 0.81±0.03                           & 0.83±0.04                           & 0.81±0.05                           & \textbf{0.85±0.02}                   & 0.83±0.01                              \\
\multicolumn{1}{r|}{}                                                                   & cPNN      & 0.79±0.03                             & 0.80±0.04                             & 0.78±0.06                             & 0.78±0.02                              & 0.79±0.01                                & 0.81±0.03                           & 0.81±0.04                           & 0.79±0.06                           & 0.79±0.02                            & 0.80±0.02                              \\
\multicolumn{1}{r|}{}                                                                   & MAcPNN    & \textbf{0.80±0.02}                    & \textbf{0.82±0.04}                    & \textbf{0.81±0.05}                    & \textbf{0.84±0.02}                     & \textbf{0.81±0.01}                       & 0.82±0.03                           & 0.83±0.04                           & \textbf{0.82±0.05}                  & 0.84±0.02                            & 0.83±0.01                              \\ \hline
\multicolumn{1}{r|}{\multirow{4}{*}{\begin{tabular}[c]{@{}r@{}}A\\ Q\end{tabular}}}     & ARF$_T$   & 0.61±0.05                             & 0.61±0.06                             & 0.60±0.06                             & 0.60±0.03                              & 0.60±0.01                                & 0.63±0.05                           & 0.63±0.05                           & 0.62±0.07                           & 0.62±0.03                            & 0.63±0.01                              \\
\multicolumn{1}{r|}{}                                                                   & cLSTM     & 0.68±0.02                             & 0.67±0.03                             & 0.64±0.02                             & 0.66±0.02                              & 0.66±0.01                                & 0.70±0.02                           & 0.69±0.03                           & 0.68±0.02                           & 0.68±0.02                            & 0.69±0.01                              \\
\multicolumn{1}{r|}{}                                                                   & cPNN      & 0.70±0.02                             & 0.70±0.02                             & 0.67±0.03                             & 0.66±0.02                              & 0.68±0.01                                & 0.70±0.02                           & 0.69±0.03                           & 0.67±0.03                           & 0.66±0.02                            & 0.68±0.01                              \\
\multicolumn{1}{r|}{}                                                                   & MAcPNN    & \textbf{0.71±0.02}                    & \textbf{0.71±0.02}                    & \textbf{0.70±0.02}                    & \textbf{0.70±0.02}                     & \textbf{0.71±0.01}                       & \textbf{0.71±0.02}                  & \textbf{0.70±0.03}                  & \textbf{0.70±0.02}                  & \textbf{0.70±0.01}                   & \textbf{0.70±0.01}                     \\ \hline
\end{tabular}

\end{table*}

\begin{figure*}[]
    \centering
    \begin{subfigure}{0.19\textwidth}
        \centering
        \includegraphics[]{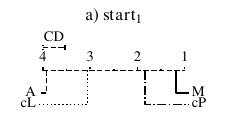}
        \phantomcaption{}
        \label{fig:nemenyi_start_1}
    \end{subfigure}
    \hfill
    \begin{subfigure}{0.19\textwidth}
        \centering
        \includegraphics[]{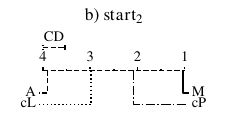}
        \phantomcaption{}
        \label{fig:nemenyi_start_2}
    \end{subfigure}
    \hfill
    \begin{subfigure}{0.19\textwidth}
        \centering
        \includegraphics[]{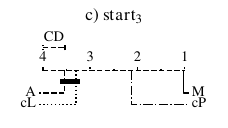}
        \phantomcaption{}
        \label{fig:nemenyi_start_3}
    \end{subfigure}
    \hfill
    \begin{subfigure}{0.19\textwidth}
        \centering
        \includegraphics[]{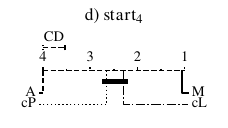}
        \phantomcaption{}
        \label{fig:nemenyi_start_4}
    \end{subfigure}
    \hfill
    \begin{subfigure}{0.19\textwidth}
        \centering
        \includegraphics[]{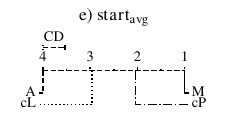}
        \phantomcaption{}
        \label{fig:nemenyi_start_avg}
    \end{subfigure}
    \hfill
    \caption{Critical distance diagram of the Nemenyi test ($\alpha=0.05$) on 100 real data streams (50 of Weather and 50 of AirQuality) on the different start$_j$ and start$_{avg}$ considering the Cohen's Kappa score. A is ARF$_T$, cL is cLSTM, M is MAcPNN, and cP is cPNN. MAcPNN is statistically the best model in all the cases. Applying MAcPNN boosts the adaptation to new concepts.}
    \label{fig:nemenyi_start}
\end{figure*}

\begin{figure*}[]
    \centering
    \begin{subfigure}{0.19\textwidth}
        \centering
        \includegraphics[]{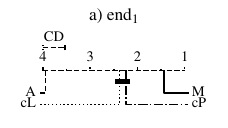}
        \phantomcaption{}
        \label{fig:nemenyi_end_1}
    \end{subfigure}
    \hfill
    \begin{subfigure}{0.19\textwidth}
        \centering
        \includegraphics[]{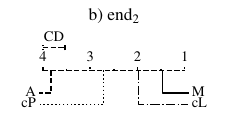}
        \phantomcaption{}
        \label{fig:nemenyi_end_2}
    \end{subfigure}
    \hfill
    \begin{subfigure}{0.19\textwidth}
        \centering
        \includegraphics[]{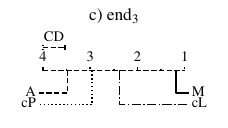}
        \phantomcaption{}
        \label{fig:nemenyi_end_3}
    \end{subfigure}
    \hfill
    \begin{subfigure}{0.19\textwidth}
        \centering
        \includegraphics[]{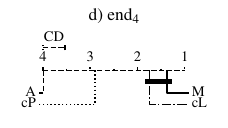}
        \phantomcaption{}
        \label{fig:nemenyi_end_4}
    \end{subfigure}
    \hfill
    \begin{subfigure}{0.19\textwidth}
        \centering
        \includegraphics[]{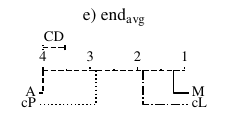}
        \phantomcaption{}
        \label{fig:nemenyi_end_avg}
    \end{subfigure}
    \hfill
    \caption{Critical distance diagram of the Nemenyi test ($\alpha=0.05$) on 100 real data streams (50 of Weather and 50 of AirQuality) on the different end$_j$ and end$_{avg}$ considering the Cohen's Kappa score. A is ARF$_T$, cL is cLSTM, M is MAcPNN, and cP is cPNN. MAcPNN is statistically the best model in all the cases (except for end$_4$ where it still has the best rank). Applying MAcPNN improves the performance of the whole concept.}
    \label{fig:nemenyi_end}
\end{figure*}

\begin{figure*}[]
    \centering
    \begin{subfigure}{0.49\textwidth}
        \centering
        \includegraphics[]{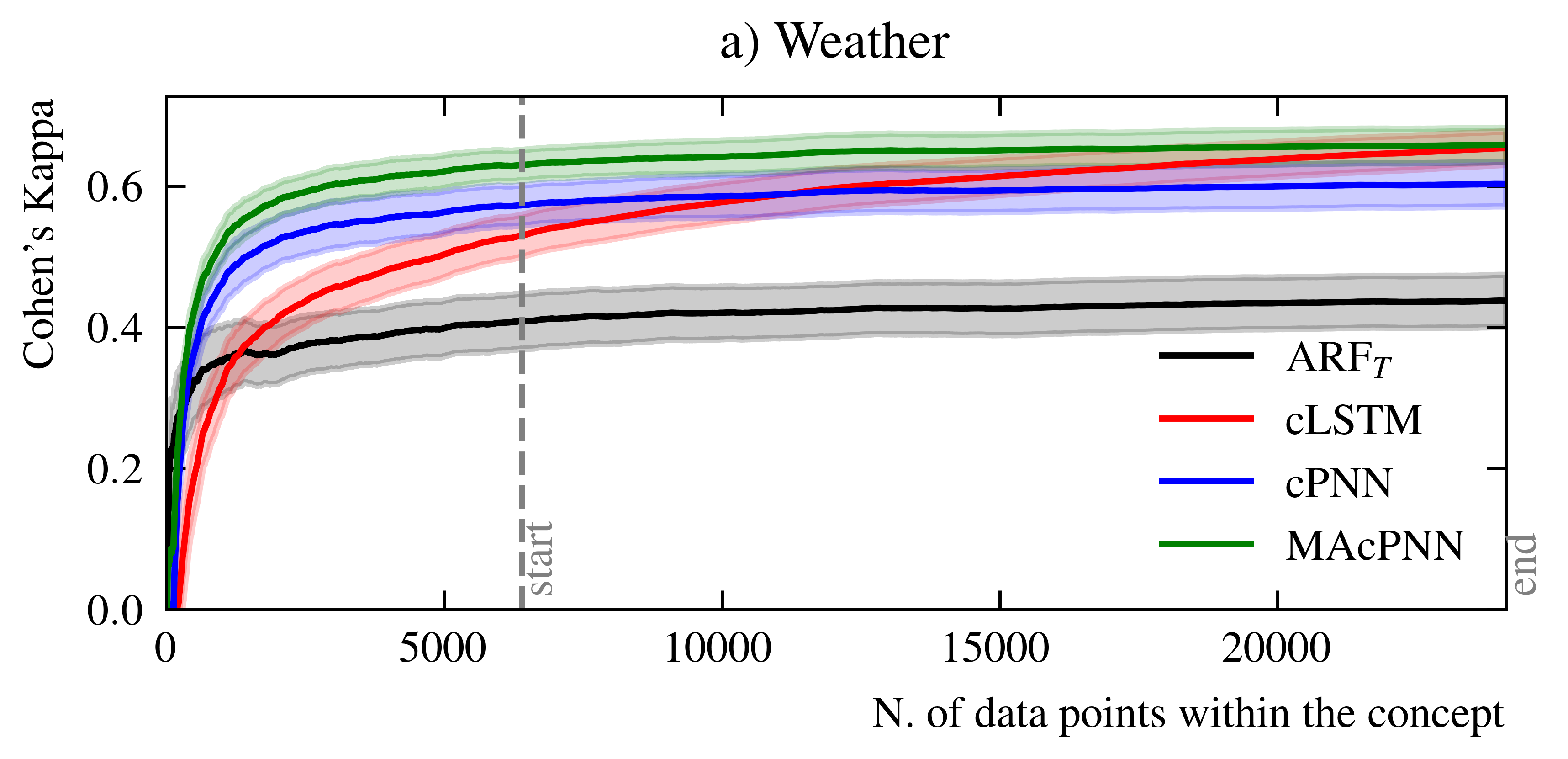}
        \phantomcaption{}
        \label{fig:performance_averaged_w}
    \end{subfigure}
    \hfill
    \begin{subfigure}{0.49\textwidth}
        \centering
        \includegraphics[]{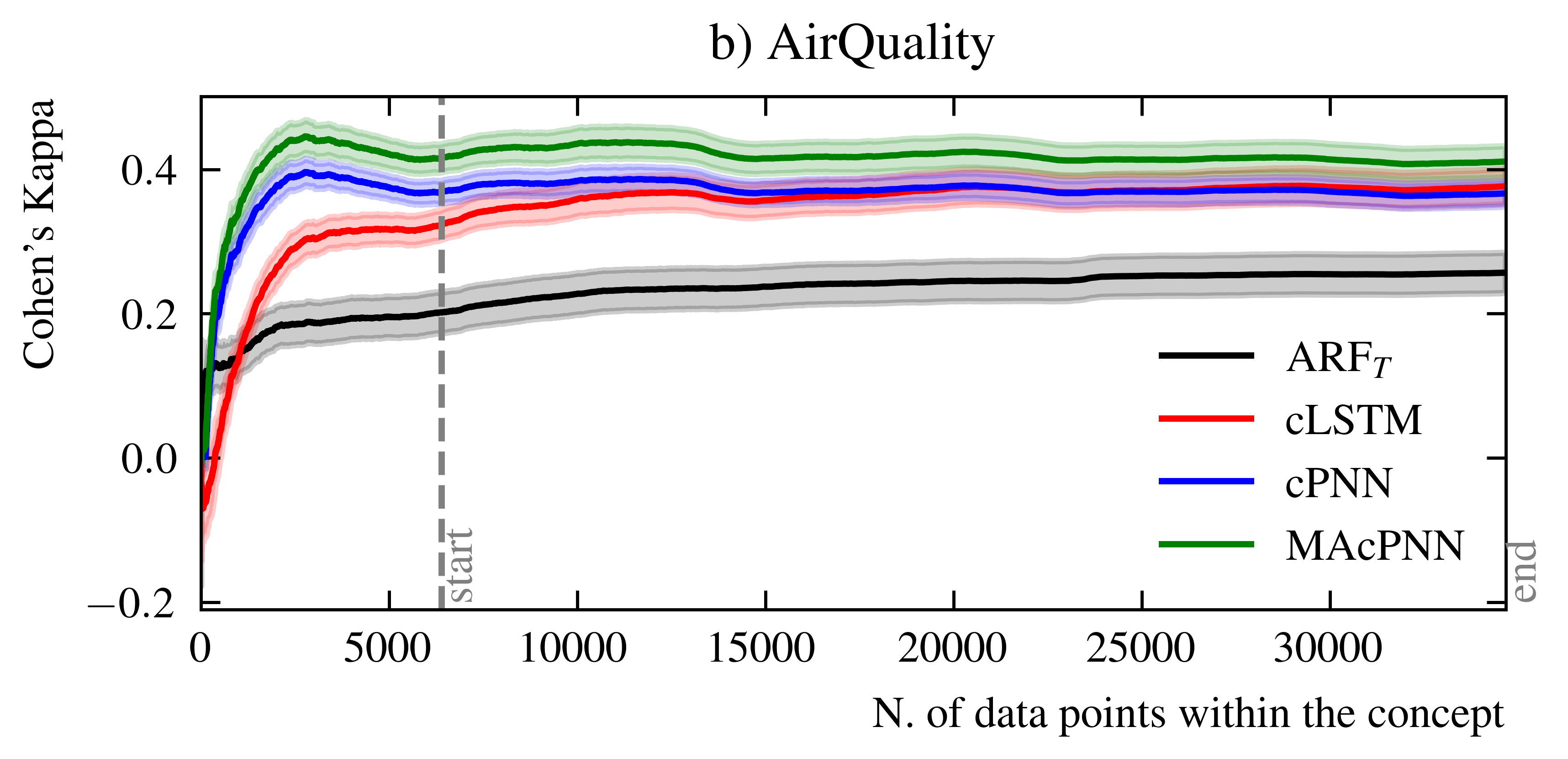}
        \phantomcaption{}
        \label{fig:performance_averaged_aq}
    \end{subfigure}
    \caption{Average Cohen's Kappa over time evolution on the network on 50 configurations of Weather (\ref{fig:performance_averaged_w}) and AirQuality (\ref{fig:performance_averaged_aq}). For each configuration, we first average the scores of the different concepts (from the second onwards) on each device separately. Then, we average the scores across the three devices, and finally, we average these scores over all configurations. MAcPNN adapts better to concept drifts. On Weather, cLSTM can approach MAcPNN performance at the end of the concept. On AirQuality, MAcPNN keeps the best models over the entire concept.}
    \label{fig:performance_averaged}
\end{figure*}

To provide a clear explanation of how the \emph{start} and \emph{end} metrics are computed, we illustrate in Fig.~\ref{fig:performance_node} the Cohen's Kappa evolution over time of Device 2 on a Weather configuration. Scores, computed through prequential evaluation, are reset after each drift. For each model, we calculate \emph{start}$_{avg}$ as the average of \emph{start$_{1}$}, ..., \emph{start$_{4}$}, and \emph{end}$_{avg}$ as the average of \emph{end$_{1}$}, ..., \emph{end$_{4}$}. The network aggregated \emph{start}$_{avg}$ and \emph{end}$_{avg}$ are computed by averaging the respective \emph{start}$_{avg}$ and \emph{end}$_{avg}$ values from the three devices. We report the best SML model. 

Tables~\ref{table:results_kappa} and \ref{table:results_acc} provide a concise overview of results obtained from 10 randomly chosen configurations of SRW and 50 randomly chosen configurations of Weather (W) and AirQuality (AQ). For each concept drift $j$, we report the average Cohen's Kappa and Balanced Accuracy scores, along with their corresponding standard deviations. Given a concept drift $j$ and a configuration, start$_j$ and end$_j$ are computed as the mean of the three devices. We average the results on all the configurations for each type of data stream. We report in bold the best-performing model. To assess whether one model outperforms another, we conduct a statistical hypothesis test with $\alpha=0.05$, considering the performance of the different configurations. We first conduct a Shapiro-Wilk test to check for normality. If we cannot reject the null hypothesis for both distributions, we conduct a Welch's t-test. Otherwise, we run a Wilcoxon signed-rank test. We perform a one-sided test in both cases. MAcPNN always outperforms all the other models during the first part of the concepts. We can, therefore, conclude that re-using the knowledge acquired by the other network devices is useful to boost the adaptation to concept drifts. cLSTM, instead, struggles to adapt to concept drifts. This situation is aligned with the findings of the original cPNN paper~\cite{cit:cpnn}. When the new concept introduces a different classification problem, cLSTM requires more adaptation steps. Additionally, the SML models cannot learn concepts incorporating elaborated temporal dependence and achieve low scores. MAcPNN always outperforms the other models at the end of the AirQuality concepts and, in most cases, SRW ones. Considering Weather's ends of concepts, cLSTM can approach MAcPNN performance. However, as evidenced by \emph{start} performance, cLSTM cannot quickly adapt to new concepts. Adopting it would mean having a model that is not reliable after a concept drift. If the concept lasts enough, cLSTM can learn it efficiently in the end, but it requires more time than MAcPNN. cPNN is never the best model. Notably, these results show that MAcPNN's mutual assistance is advantageous. However, during temporary disconnections, local models can still learn, albeit more slowly.

To confirm our findings on the real datasets, we conduct the Nemenyi posthoc test ($\alpha=0.05$)~\cite{cit:nemenyi} to measure the true difference between models' Cohen's Kappa scores by constructing the critical difference (CD) diagram using the average ranks. The axis represents the average rank, while CD is the minimum difference between the two ranks that must be considered statistically different. We consider the 100 real data streams (50 randomly selected configurations of Weather and 50 of AirQuality). Figure~\ref{fig:nemenyi_start} reports the different start$_j$ and start$_{avg}$ results. Figure~\ref{fig:nemenyi_end} shows the same analyses on the different end$_j$ and end$_{avg}$. MAcPNN demonstrates statistically the best model in all the cases during the first part of the concepts. The same findings emerge from the analyses of the performance of the whole concepts (except for the last drift, where it still has the best rank but no statistical difference). These tests confirm that MAcPNN boosts the adaptation to new concepts and improves performance when considering the whole concept. Finally, Fig.~\ref{fig:performance_averaged} illustrates Cohen's Kappa evolution on the network. We compute the Cohen's Kappa scores for each model, device, and configuration for each concept (the first concept is excluded). Then, we average the scores on the different concepts for each model and device and the scores of the three network devices. Finally, we average the scores of all the configurations. We represent the distribution spread at each point by plotting the standard deviation over time. The findings confirm MAcPNN as the best initial model. cLSTM approaches MAcPNN performance by the end of the Weather concepts but adapts slowly to drifts.

Concerning the number of communications, we have three devices ($n=3$), each with four drifts. The total number of communications is \mbox{$C_{ours} = n \cdot 4 \cdot (n-1) = 3 \cdot 4 \cdot 2 = 24$}. SRW's number of mini-batches $N_B$ is around 977, Weather's is around 942, and AirQuality's is around 1355. The number of communications in the naive approach is \mbox{$C_{naive} = 3 \cdot N_B \cdot (n-1)$}, which results in 5862, 5652, and 8130, respectively. Our number of communications is $0.3\%-0.4\%$ of the naive's one.

\section{Conclusion} \label{sec:conclusion}
This work proposes Mutual Assisted Learning (MAL), a new learning paradigm to connect edge machine learning models in a network where different users' devices are trained on specific data streams, as in the Internet of Things. The assumption is that each device observes different concepts, and concept drifts are not synchronous. The goal is to reduce the number of communications between the devices while allowing each device to ask for assistance from the others when encountering a new problem. This is done without using a central orchestrator. Each device is autonomous. We base our research on Vygotsky's popular Sociocultural Theory of Cognitive Development to do so. This theory is grounded in the Zone of Proximal Development, where learning is enhanced with individuals' assistance. In our network, edge devices act as individuals. When a device detects concept drift, it can request assistance from others and then decide to use the assistance or continue learning independently. This paradigm overcomes the limits of Federated Learning, which requires communication between devices at each training round and aims to aggregate the knowledge for a common general problem. A crucial requirement is to enable each single device to jointly address the problems of continuous learning, concept drifts, temporal dependence, and catastrophic forgetting. In this scenario, avoiding forgetting gains more importance since a device may benefit from using others' past knowledge. We build on Continuous Progressive Neural Networks and introduce MAcPNN, an approach using MAL in a network of cPNNs. We present a new loss function to obtain anytime classifiers able to make predictions on a single data point. Since each device could have limited resources and we also need to transfer the models through the network, we propose a solution to reduce the cPNN memory footprint using quantization. Notably, quantization also makes building an ensemble of models less expensive. We reduce the number of communications by making devices communicate only after concept drifts. Our experimentation compares MAcPNN against SML and cPNN models trained only on the data stream associated with each specific device. MAcPNN outperforms all the other models.

During the experimentation, we assumed a concept drift detector with 100\% accuracy and focused on measuring the adaptation to drifts. A crucial future work is to blend the cPNN architecture with a real concept drift detector. Since cPNN is robust by construction to forgetting, we focus on the prequential evaluation. Measuring forgetting in SML and cLSTM could highlight cPNN's ability to retain past knowledge. Hyperparameter tuning remains another open issue. Additionally, the proposed solution is tested only on abrupt concept drifts. Regarding MAcPNN, the current solution is for a network with few devices. We intend to test MAcPNN in larger networks and connect each device only with its neighbors to avoid the quadratic growth of communications. To select ensemble models, we prioritize those with higher timestamps. Alternative criteria based on performance may also be explored.

\bibliographystyle{splncs04}
\bibliography{bibliography}

@article{cit:catastrophic_forgetting,
  author    = {Matthias De Lange and
               Rahaf Aljundi and
               Marc Masana and
               Sarah Parisot and
               Xu Jia and
               Ales Leonardis and
               Gregory G. Slabaugh and
               Tinne Tuytelaars},
  title     = {{A Continual Learning Survey: Defying Forgetting in Classification
               Tasks}},
  journal   = {{IEEE} Trans. Pattern Anal. Mach. Intell.},
  volume    = {44},
  number    = {7},
  pages     = {3366--3385},
  year      = {2022}
}

@incollection{cit:stability_plasticity,
  title={{Catastrophic interference in connectionist networks: The sequential learning problem}},
  author={McCloskey, Michael and Cohen, Neal J},
  booktitle={Psychology of learning and motivation},
  volume={24},
  pages={109--165},
  year={1989},
  publisher={Elsevier}
}

@book{cit:deep_learning,
  author    = {Ian J. Goodfellow and
               Yoshua Bengio and
               Aaron C. Courville},
  title     = {{Deep Learning}},
  series    = {Adaptive computation and machine learning},
  publisher = {{MIT} Press},
  year      = {2016}
}

@inproceedings{cit:gama_learning_with_dd,
  author    = {Jo{\~{a}}o Gama and
               Pedro Medas and
               Gladys Castillo and
               Pedro Pereira Rodrigues},
  title     = {{Learning with Drift Detection}},
  booktitle = {{SBIA}},
  series    = {LNCS},
  volume    = {3171},
  pages     = {286--295},
  publisher = {Springer},
  year      = {2004}
}

@article{cit:lstm,
  author    = {Sepp Hochreiter and
               J{\"{u}}rgen Schmidhuber},
  title     = {{Long Short-Term Memory}},
  journal   = {Neural Comput.},
  volume    = {9},
  number    = {8},
  pages     = {1735--1780},
  year      = {1997}
}

@inproceedings{cit:gim,
  author    = {Andrea Cossu and
               Antonio Carta and
               Davide Bacciu},
  title     = {{Continual Learning with Gated Incremental Memories for sequential
               data processing}},
  booktitle = {{IJCNN}},
  pages     = {1--8},
  publisher = {{IEEE}},
  year      = {2020}
}

@article{cit:ilstm,
  title={{An Incremental Learning Approach Using Long Short-Term Memory Neural Networks}},
  author={Lemos Neto, {\'A}lvaro C and Coelho, Rodrigo A and Castro, Cristiano L de},
  journal={Journal of Control, Automation and Electrical Systems},
  pages={1--9},
  year={2022},
  publisher={Springer}
}

@article{cit:concept_drift,
  author    = {Jie Lu and
               Anjin Liu and
               Fan Dong and
               Feng Gu and
               Jo{\~{a}}o Gama and
               Guangquan Zhang},
  title     = {{Learning under Concept Drift: {A} Review}},
  journal   = {{IEEE} Trans. Knowl. Data Eng.},
  volume    = {31},
  number    = {12},
  pages     = {2346--2363},
  year      = {2019}
}

@article{cit:cl,
  author    = {Timoth{\'{e}}e Lesort and
               Vincenzo Lomonaco and
               Andrei Stoian and
               Davide Maltoni and
               David Filliat and
               Natalia D{\'{\i}}az Rodr{\'{\i}}guez},
  title     = {Continual learning for robotics: Definition, framework, learning strategies,
               opportunities and challenges},
  journal   = {Inf. Fusion},
  volume    = {58},
  pages     = {52--68},
  year      = {2020}
}

@inproceedings{cit:prequential,
  author    = {Jo{\~{a}}o Gama and
               Raquel Sebasti{\~{a}}o and
               Pedro Pereira Rodrigues},
  title     = {{Issues in evaluation of stream learning algorithms}},
  booktitle = {{KDD}},
  pages     = {329--338},
  publisher = {{ACM}},
  year      = {2009}
}

@book{book_bifet,
  title={{Machine learning for data streams: with practical examples in MOA}},
  author={Bifet, Albert and Gavald{\`a}, Ricard and Holmes, Geoff and Pfahringer, Bernhard},
  year={2018},
  publisher={MIT press}
}

@article{cit:streaming_ts_zliobaite,
  author    = {Indre Zliobaite and Albert Bifet and
               Jesse Read and
               Bernhard Pfahringer and
               Geoff Holmes},
  title     = {{Evaluation methods and decision theory for classification of streaming data with temporal dependence}},
  journal   = {Mach. Learn.},
  volume    = {98},
  number    = {3},
  pages     = {455--482},
  year      = {2015}
}

@article{Ziffer22,
  title={{Towards time-evolving analytics: Online learning for time-dependent evolving data streams}},
  author={Ziffer, Giacomo and Bernardo, Alessio and {Della Valle}, Emanuele and Cerqueira, Vitor and Bifet, Albert},
  journal={Data Science},
  year={2023},
  volume={6},
  number={1-2},
  pages={1--16},
  publisher={IOS Press}
}

@inproceedings{ADWIN,
  author    = {Albert Bifet and
               Ricard Gavald{\`{a}}},
  title     = {{Learning from Time-Changing Data with Adaptive Windowing}},
  booktitle = {{SDM}},
  pages     = {443--448},
  publisher = {{SIAM}},
  year      = {2007}
}

@article{ARF,
  author    = {Heitor Murilo Gomes and
               Albert Bifet and
               Jesse Read and
               Jean Paul Barddal and
               Fabr{\'{\i}}cio Enembreck and
               Bernhard Pfahringer and
               Geoff Holmes and
               Talel Abdessalem},
  title     = {{Adaptive random forests for evolving data stream classification}},
  journal   = {Mach. Learn.},
  volume    = {106},
  number    = {9-10},
  pages     = {1469--1495},
  year      = {2017}
}

@inproceedings{DBLP:conf/sdm/BifetG07,
  author    = {Albert Bifet and
               Ricard Gavald{\`{a}}},
  title     = {{Adaptive Learning from Evolving Data Streams}},
  booktitle = {{IDA}},
  series    = {LNCS},
  volume    = {5772},
  pages     = {249--260},
  publisher = {Springer},
  year      = {2009}
}

@inproceedings{cit:cpnn,
  author       = {Giannini, Federico and
                  Ziffer, Giacomo and
                  Della Valle, Emanuele},
  title        = {{cPNN: Continuous Progressive Neural Networks for Evolving Streaming
                  Time Series}},
  booktitle    = {{PAKDD} {(4)}},
  series       = {Lecture Notes in Computer Science},
  volume       = {13938},
  pages        = {328--340},
  publisher    = {Springer},
  year         = {2023}
}

@inproceedings{cit:fedearated,
  author       = {Brendan McMahan and
                  Eider Moore and
                  Daniel Ramage and
                  Seth Hampson and
                  Blaise Ag{\"{u}}era y Arcas},
  title        = {Communication-Efficient Learning of Deep Networks from Decentralized
                  Data},
  booktitle    = {{AISTATS}},
  series       = {Proceedings of Machine Learning Research},
  volume       = {54},
  pages        = {1273--1282},
  publisher    = {{PMLR}},
  year         = {2017}
}

@article{cit:quantization,
  author       = {Tailin Liang and
                  John Glossner and
                  Lei Wang and
                  Shaobo Shi and
                  Xiaotong Zhang},
  title        = {{Pruning and quantization for deep neural network acceleration: A
                  survey}},
  journal      = {Neurocomputing},
  volume       = {461},
  pages        = {370--403},
  year         = {2021}
}

@book{cit:tsa,
  title={{Time series analysis: forecasting and control}},
  author={Box, George EP and Jenkins, Gwilym M and Reinsel, Gregory C and Ljung, Greta M},
  year={2015},
  publisher={John Wiley \& Sons}
}

@article{cit:makridakis_m5,
  title={{M5 accuracy competition: Results, findings, and conclusions}},
  author={Makridakis, Spyros and Spiliotis, Evangelos and Assimakopoulos, Vassilios},
  journal={IJOF},
  volume={38},
  number={4},
  pages={1346--1364},
  year={2022},
  publisher={Elsevier}
}

@inproceedings{cit:edge_computing,
  author       = {Tyler Holmes and
                  Charlie McLarty and
                  Yong Shi and
                  Patrick Bobbie and
                  Kun Suo},
  title        = {{Energy Efficiency on Edge Computing: Challenges and Vision}},
  booktitle    = {{IPCCC}},
  pages        = {1--6},
  publisher    = {{IEEE}},
  year         = {2022}
}

@inproceedings{cit:fl_ds2,
  author       = {Yujing Chen and
                  Zheng Chai and
                  Yue Cheng and
                  Huzefa Rangwala},
  title        = {{Asynchronous Federated Learning for Sensor Data with Concept Drift}},
  booktitle    = {{IEEE} BigData},
  pages        = {4822--4831},
  publisher    = {{IEEE}},
  year         = {2021}
}

@article{cit:fl_ds_fedstream,
  author       = {Cobbinah Bernard Mawuli and
                  Liwei Che and
                  Jay Kumar and
                  Salah Ud Din and
                  Zhili Qin and
                  Qinli Yang and
                  Junming Shao},
  title        = {{FedStream: Prototype-Based Federated Learning on Distributed Concept-Drifting
                  Data Streams}},
  journal      = {{IEEE} Trans. Syst. Man Cybern. Syst.},
  volume       = {53},
  number       = {11},
  pages        = {7112--7124},
  year         = {2023}
}

@article{cit:fl_survey,
  author       = {Chen Zhang and
                  Yu Xie and
                  Hang Bai and
                  Bin Yu and
                  Weihong Li and
                  Yuan Gao},
  title        = {A survey on federated learning},
  journal      = {Knowl. Based Syst.},
  volume       = {216},
  pages        = {106775},
  year         = {2021}
}

@article{cit:p2p_fl1,
  author       = {Anusha Lalitha and
                  Osman Cihan Kilinc and
                  Tara Javidi and
                  Farinaz Koushanfar},
  title        = {{Peer-to-peer Federated Learning on Graphs}},
  journal      = {CoRR},
  volume       = {abs/1901.11173},
  year         = {2019}
}

@article{cit:p2p_fl2,
  author       = {Abhijit Guha Roy and
                  Shayan Siddiqui and
                  Sebastian P{\"{o}}lsterl and
                  Nassir Navab and
                  Christian Wachinger},
  title        = {{BrainTorrent: {A} Peer-to-Peer Environment for Decentralized Federated
                  Learning}},
  journal      = {CoRR},
  volume       = {abs/1905.06731},
  year         = {2019}
}

@inproceedings{cit:p2p_fl3,
  author       = {Tobias Wink and
                  Zolt{\'{a}}n Nochta},
  title        = {{An Approach for Peer-to-Peer Federated Learning}},
  booktitle    = {{DSN} Workshops},
  pages        = {150--157},
  publisher    = {{IEEE}},
  year         = {2021}
}

@article{cit:nemenyi,
  author       = {Janez Demsar},
  title        = {{Statistical Comparisons of Classifiers over Multiple Data Sets}},
  journal      = {J. Mach. Learn. Res.},
  volume       = {7},
  pages        = {1--30},
  year         = {2006}
}

@article{cit:cohen_kappa,
  title={{Interrater reliability: the kappa statistic}},
  author={McHugh, Mary L},
  journal={Biochemia medica},
  volume={22},
  number={3},
  pages={276--282},
  year={2012},
  publisher={Medicinska naklada}
}

@article{cit:vygotsky,
  title={Vygotsky's sociocultural theory and contributions to the development of constructivist curricula},
  author={Jaramillo, James A},
  journal={Education},
  volume={117},
  number={1},
  pages={133--141},
  year={1996},
  publisher={Project Innovation Austin LLC}
}

@article{cit:weather,
  title={{Eleven years of mountain weather, snow, soil moisture and streamflow data from the rain--snow transition zone--the Johnston Draw catchment, Reynolds Creek Experimental Watershed and Critical Zone Observatory, USA}},
shorttitle={Eleven years of mountain weather data},
  author={Godsey, Sarah E and others},
  journal={Earth System Science Data},
  volume={10},
  number={3},
  pages={1207--1216},
  year={2018},
  publisher={Copernicus GmbH}
}

\end{document}